# CARRS: A Constraint-Aware, Real-Time Routing System for Emergency Vehicles in Urban Environments

VINCENT, V, SELLO

Academy of Computer Science and Software Engineering, University of Johannesburg, Johannesburg, Gauteng, South Africa, vincentiussello@gmail.com

TIMOTHY, T, ADELIYI

Department of Informatics, University of Pretoria, Pretoria, Gauteng, South Africa, timothy.adeliyi@up.ac.za

RIDEWAAN, R, HANSLO

Department of Information Systems, University of Cape Town, Cape Town, Western Cape, South Africa, ridewaan.hanslo@uct.ac.za

Emergency vehicles such as police vehicles, fire engines, and ambulances require a stable and efficient route optimization to reach life-critical events in the shortest times possible. However, the dynamic nature of urban traffic makes this ideal scenario much more challenging, causing critical life-threatening delays. Traditional civilian navigation systems cannot account for real-time traffic changes alongside vehicle-specific constraints, creating inadequate routing scenarios. This research presents the Constraint-Aware Real-Time Routing System (CARRS), a unified software architecture tailored for emergency coordination. The CARRS architecture is engineered to fuse together classical pathfinding algorithms such as A*, Dijkstra, and Bidirectional A*, with Constraint Satisfaction Problem methodologies. The Design Science Research Methodology is the research framework, and prototyping is the iterative process. This research strategy guides the investigation, producing a system that dynamically computes obstacle-aware, legally compliant routes for emergency vehicles. The artifact was empirically evaluated using OpenStreetMap data for major metropolitan road networks, combined with simulated dynamic traffic scenarios. Performance metrics focused on route optimality, constraint adherence, rerouting responsiveness, and the potential for future system scalability. Findings revealed the system's capability to generate optimal routes for operationally relevant subgraphs in under 35 seconds, with incremental rerouting responsiveness below 20 seconds. Furthermore, the system achieved more than 90% adherence to vehicle-specific constraints, making the CARRS architecture viable for real-time dynamic routing requirements. This study delivers a routing platform that bridges artificial intelligence-driven optimization with software engineering operational compliance, presenting a practical transformative step toward adaptive emergency vehicle coordination in dynamic urban environments.

## 1 INTRODUCTION

Efficient emergency routing saves lives and property, making every second critical [1]. However, first responders navigating urban environments face a chaotic mix of dynamic obstacles, such as traffic congestion and accidents, and strict physical constraints, including vehicle weight limits, turning radius limitations, and access permissions [2, 3]. Traditional navigation systems simply aren't adequately equipped to juggle these real-time and static variables simultaneously [4].

Constraint Satisfaction Problem (CSP) methodologies offer a structured way to model these complex operational rules [5, 6]. Albeit, because CSPs are computationally heavy (often classified as NP-complete), solving them by brute force during an active emergency is impractical [7]. While advanced search strategies and heuristic filtering help navigate these complexities [8 - 10], most existing routing systems still fall short. For example, they typically optimize for travel time under simplified or static conditions [11]. In addition, when systems do handle dynamic updates, they often validate constraints after finding a path. This disconnected approach generates unsafe or legally invalid routes, forcing dispatchers to intervene manually and waste precious time.

This operational gap highlights this study's research question: *How can we design a routing system that computes real-time, mathematically optimal, and strictly constraint-compliant routes for emergency vehicles in dynamic cities?* To answer this, we introduce the Constraint-Aware Real-Time Routing System (CARRS). Our specific contributions include:

- **Integrating** classical shortest-path algorithms (A*, Dijkstra, Bidirectional A*) directly with CSP-based modelling to guarantee both route optimality and operational adherence.
- **Developing** a multi-threaded software architecture to ensure dispatcher interfaces remain highly responsive during intense computational loads.
- **Evaluating** the system's constraint adherence and rerouting speeds under simulated dynamic traffic conditions.

By embedding CSP principles directly into the pathfinding loop, the CARRS architecture eliminates the gap between theoretical routing and real-world compliance. This unified software engineering framework enhances dispatcher situational awareness, reduces manual routing delays, and is posited to support the broader societal goal of resilient and intelligent emergency response.

## 2 LITERATURE REVIEW

This section examines previous research across classical shortest path algorithms, constraint satisfaction approaches, real-time routing techniques, and specialized emergency vehicle systems, articulating the exact gaps our research addresses.

**Classical Algorithms and their Limits.** Shortest path algorithms are the engine of modern navigation. While Dijkstra's foundational algorithm guarantees optimal routes, it explores too much of the graph to be practical in dense, large-scale cities [13, 14]. The A* algorithm solved this computational bottleneck by introducing heuristic guidance, drastically reducing unnecessary exploration while preserving route optimality [15–17]. Although bidirectional A* search variants attempt to improve this further [18, 19], they often stumble in the irregular layouts of real-world metropolitan grids [20], which is core to this investigation. More critically, none of these classical algorithms inherently understand complex operational rules, such as bridge weight limits, tunnel clearances, or emergency-only access lanes.

**Constraint Satisfaction and Real-Time Routing.** To bridge this gap, Constraint Satisfaction Problems (CSPs) offer a structured way to encode physical and legal vehicle restrictions [21, 22]. Research by Riera et al. [23] demonstrated that CSPs provide the exact modularity needed to juggle diverse vehicle profiles, such as fire engines, ambulances, and police vehicles. Modern approaches are finding that integrating this CSP filtering directly into the heuristic search process, thereby, dynamically validating constraints during node expansion rather than post-processing, which drastically improves efficiency and responsiveness in dynamic traffic environments [24, 25].

**Subgraphs and Emergency Systems.** Even with integrated algorithms, exhaustively searching an entire city grid in real time forces a painful trade-off between optimality and speed [26]. Subgraph-based routing solves this problem by extracting only the geographically relevant network region [27], an approach uniquely suited for the rapid decision-making required in emergencies [28]. Despite these advances, many emergency systems still rely on civilian navigation infrastructures that fail to support specialized vehicle constraints [29]. Developers are often forced to compromise, for example, by prioritizing routing speed at the expense of legal compliance or enforce strict rules and suffer crippling computational overhead [30, 31].

The literature consistently concludes that the most effective routing systems validate constraints directly within the routing loop [32]. Building on this consensus, our research delivers a unified software engineering framework that embeds CSP reasoning directly into A* search expansion. By pairing this with subgraph extraction and multi-threading, the CARRS architecture effectively solves the scalability and operational gaps in current urban emergency navigation.

## 3 METHODOLOGY AND IMPLEMENTATION

This research adopts the Design Science Research Methodology (DSRM) as its guiding framework. DSRM provides a structured, iterative process for developing, evaluating, and refining technological artifacts aimed at solving real-world problems [33-34]. In the context of this research, the artifact is CARRS, supporting optimal decision making for emergency vehicle routing in dynamic urban environments. The DSRM iterative process involves six key stages, these stages ensure the quality and reliability of the study. This methodological approach ensures not only a technical thorough development cycle but also empirical validation through performance metrics and scenario-based testing.

i. **Problem Identification and Motivation.** This study begins by identifying inefficiencies in existing emergency routing systems that lack real-time adaptability and constraint enforcement.

ii. **Objective Definition.** Derived from the problem's context, the design objectives involve developing an artifact capable of integrating real-time simulated traffic data with classical pathfinding and constraint satisfaction algorithms. Furthermore, the project establishes measurable goals, namely: real-time performance, constraint adherence, rapid rerouting, and scalability.

iii. **Design and Development.** The artifact is developed through iterative prototyping. In Design Science Research (DSR), prototyping is a critical iterative method used to rapidly evaluate design alternatives [34]. Furthermore, the instantiation of a prototype allows researchers to uncover systemic risks, such as performance bottlenecks and integration challenges, early in the development lifecycle rather than during final implementation [35-36]. In essence, DSR is the research strategy or framework, and prototyping is the iterative practical implementation method or process. This iterative process allows for continuous refinement of the artifact based on insights gained during experimentation and feedback from simulated test scenarios.

iv. **Demonstration.** The artifact is validated through simulation using real world datasets from South African metropolitan road networks, with simulated traffic data. These simulations help visualize how the system responds to emergencies and dynamically reroutes based on traffic input while adhering to all constraint put in place.

v. **Evaluation.** The system's performance is assessed against defined criteria (metrics), specifically route optimality, constraint adherence, runtime, and scalability.

vi. **Communication.** This research concludes by documenting the construction of the proposed software artifact and presenting its operational outcomes. This final phase emphasizes the design rationale, architectural decisions, and implications for future research and practical adoption.

**System Architecture.** The architecture follows a modular, layered pattern to ensure separation of concerns and scalability. As illustrated in Figure 1, the system consists of five primary interacting layers:

i. **Input and Data Layer.** Imports OpenStreetMap (OSM) and simulated traffic data, and loads vehicle profiles defining height, weight, lane permissions, tunnel/bridge compatibility, and emergency privileges.

ii. **Preprocessing Layer.** Converts OSM data into weighted directed graphs using OSMnx and NetworkX, indexing road attributes and constructing subgraphs for targeted routing.

iii. **Core Routing Engine.** Implements A*, Dijkstra, and Bidirectional A* with optimizations such as Euclidean heuristic tuning and reduced memory overhead.

iv. **Constraint Satisfaction Module.** Defines constraints as a CSP model and applies per-edge validation, route-level validation, and progressive relaxation with penalty accumulation to retain feasibility when strict constraints block all paths.

v. **Visualization and UI Layer.** Displays routes, traffic conditions, constraint violations, and rerouting events with real-time responsiveness powered by background threads.

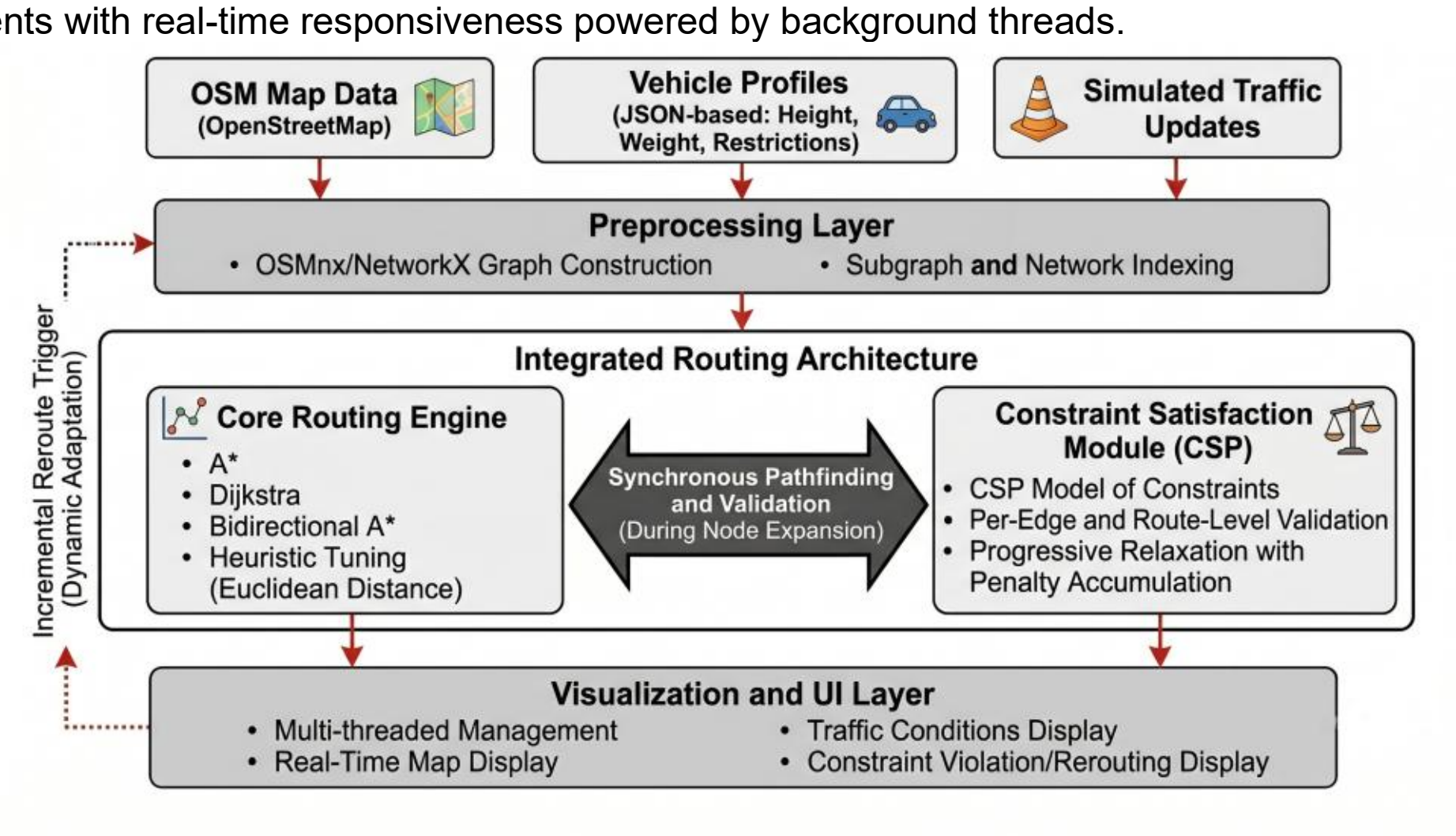

Figure 1: Constraint-Aware Real-Time Routing System (CARRS) Architecture

**Constraint Modelling Definition.** Formally, the route constraint model was defined as:

$$CM = (E, A, P)$$

Where:

- **E:** A set of edges or road segments in the graph.
- **A:** A set of attributes associated with each edge (such as lane requirements, road-closures, bridges, tunnels, and highways).
- **P:** A set of vehicle specific constraint profiles, determining the allowed conditions and penalties (e.g., fire engine, ambulance, and police vehicle).

**Performance Optimization and Evaluation Metrics.** To ensure computational efficiency and usability, several optimizations were implemented:

- **Graph subset optimization.** Restrict computations to relevant geographic regions rather than the entire metropolitan graph, significantly reducing runtime and improving scalability.
- **Heuristic tuning.** Applies consistent Euclidean distance scaling factors to optimize A* heuristic admissibility and efficiency.
- **Multi-threading.** Utilizes background processing threads to ensure the interface remains responsive during heavy computation.
- **Memory management.** Implements object pooling and garbage collection tuning to reduce memory overhead.
- **Error handling and logging.** Integrates mechanisms to record computation progress, constraint violations, and reroute events to enhance post-analysis and debugging.

To rigorously assess the system's operational viability, performance was evaluated against four primary metrics (Table 1).

Table 1: CARRS Evaluation Metrics

| Metric type | Description | Target |
|---|---|---|
| Constraint adherence | Percentage of route fully satisfying vehicle specific constraints | ≥ 90% |
| Rerouting responsiveness | Time to recompute route after event trigger | $< 20\ s$ |
| Scalability | Runtime vs graph size | Real-time feasibility |
| Computational efficiency | Average runtime and memory use | Lower is better (variable) |

## 4 RESULTS AND DISCUSSION

**Algorithmic Performance.** These results in Table 2 show that A* consistently outperformed Dijkstra and Bidirectional A* across all test cases. The heuristic driven approach of A* effectively reduced node expansion by 72% compared to Dijkstra and by 90% compared to Bidirectional A*, yielding faster route computations without sacrificing optimality. Although Bidirectional A* is theoretically faster for certain problems, it underperformed in our study due to heuristic asymmetry and non-Euclidean nature of urban road networks. A runtime of under one second (s) on subgraph scenarios demonstrates that A* is suitable for real-time decision making [29], satisfying the system’s target of near-instantaneous route generation.

Table 2: CARRS Algorithmic Performance

| Algorithm | Average Runtime (s) | Nodes Expanded | Memory Usage (mb) |
|---|---|---|---|
| A* | 0.948 | 998 | 24.12 |
| Dijkstra | 1.261 | 3 612 | 24.54 |
| Bidirectional A* | 2.404 | 10 249 | 24.82 |

**Scalability Evaluation.** The subset graph results demonstrate practical feasibility, meeting the < 35s target (Table 3). However, the full city graph exceeds 24 minutes, highlighting scalability challenges common in large graph search problems. These findings validate the architecture choice to implement graph subset optimization for operational deployment, focusing computational resources on the relevant geographic region rather than the entire metropolitan area. This aligns with the literature findings that large scale shortest path computation requires either hierarchical or region-based subgraph optimization to remain computationally viable during live emergency operations [9, 12].

Table 3: CARRS Scalability Evaluation

| Graph | Nodes | Edges | Runtime (s) | Memory Usage (mb) | Practicality |
|---|---|---|---|---|---|
| Subset Area | 1 827 | 5 133 | 32.66 | 63.91 | Viable |
| Full City | 81 807 | 208 214 | 1 476.44 | 2 187.79 | Challenging |

**Constraint Handling Performance.** Constraint satisfaction was evaluated throughout the three emergency vehicle profiles (ambulance, fire engine, police vehicles). The system's ability to maintain legal and operational validity was measured using a constraint adherence rate. Overall, the system achieved 90-100% adherence across all test cases. The minor deviations observed in fire engine tunnel access compliance stem from constraint relaxation applied to maintain feasible routing under extreme blockage conditions. The progressive relaxation strategy ensured that when strict constraint enforcement prevented route feasibility, controlled penalties were applied to allow operational viable alternatives. The balance between strict compliance and practicality, proved important in maintaining uninterrupted service continuity during route failures or blockages.

Testing under four relaxation levels namely, 0 (strict), 1 (moderate), 2 (lenient), and 3 (minimal) revealed predictable performance behaviour. The relaxation-level analysis revealed that all tested relaxation modes preserved route validity and did not significantly alter computational runtime. This stability confirmed that the CSP-based constraint engine integrates smoothly with the pathfinding layer, providing both enforceability when possible and graceful degradation when necessary. To contextualize how the system achieves this balance between strict compliance and progressive relaxation, Figure 2 outlines the per-edge validation used for the CARRS architecture. The flowchart demonstrates how the system separates hard physical limits (which trigger route failure) from soft operational constraints (which accumulate computational penalties). In essence, under severe blockage situations, the system may reliably degrade and discover operationally viable alternatives due to this penalty-accumulation architecture, without compromising vehicle safety.

**Dynamic Replanning and Rerouting.** To evaluate real-time adaptability, artificial traffic events were introduced to force rerouting in the middle of the operation. To visually demonstrate this adaptability, Figure 3 illustrates a complex rerouting scenario within the Johannesburg metropolitan network. As shown, the system initially plotted a primary path (red) but successfully generated consecutive operational alternatives (Reroute 1 (blue) and Reroute 2 (orange)) in response to dynamic traffic events along the way. The less than 30s average response time validates the system's multi-threaded design and incremental rerouting strategy. The success rate of 98% across scenarios indicates that even under continuous route updates, the system was able to maintain path optimality and interface responsiveness. Rerouting always occurs seamlessly without the user interface freezing or data loss.

**User Experience and Interpretability.** User experience testing confirmed that system refinement such as colour coded route history, vehicle simulation, and traffic legends (as depicted in the interface of Figure 3) enhance interpretability for emergency coordinators. The multi-threaded architecture prevented interface locking during heavy computation, a common challenge in routing software. Qualitative user feedback highlighted improved situational awareness and confidence in the routing recommendations.

**Discussion of findings.** The findings confirm that CARRS achieves a balanced synthesis of algorithmic performance, constraint compliance, and usability. A* consistently outperforms Dijkstra and Bidirectional A* in runtime and node expansion, aligning with theoretical expectations and empirical results in AI pathfinding research [18]. With regards to constraint awareness, the integration of CSP ensures route legality without compromising efficiency, addressing a core limitation of conventional routing systems [29]. The results indicated a less than 30s average rerouting responsiveness. In urban traffic optimization, sub-minute routing updates are classified as real-time viable, demonstrating operational feasibility in high-pressure emergency environments [28, 29]. As depicted in the findings, there are scalability trade-offs, in other words, applying a graph subset optimization strategy mitigates scalability issues effectively when compared to a full-scale graph (map) implementation. Lastly, for operational usability; multi-threading, visualization, and error-handling refinements transform a theoretical prototype into a potential real-time emergency routing application. Overall, the system meets its defined performance targets and provides a solid foundation for deployment in smart city emergency management infrastructures.

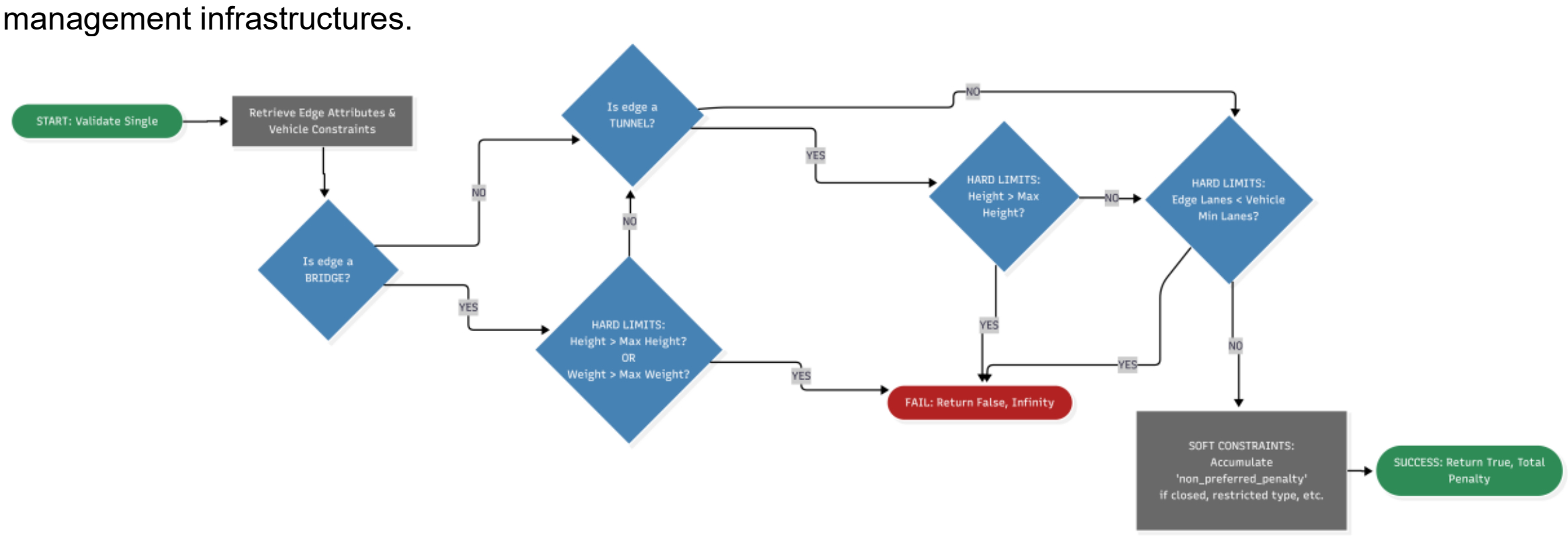

Figure 2: Algorithmic decision flow for per-edge constraint validation, illustrating the progression from hard physical limit checks (infeasible routes) to soft operational constraint checks (penalty accumulation)

**Practical implications.** Real-time constraint-aware routing ensures that ambulances, fire engines, and police vehicles consistently follow feasible, legal, and obstruction-free routes. The demonstrated computation times and rerouting responsiveness suggest substantial reductions in operational delays during emergencies, which corroborates previous studies that link AI-driven dynamic routing to significant decreases in emergency vehicle transit times [28, 29]. The system's real-time feedback, constraint indicators, and clear visualizations allow dispatchers to rely on automated route decisions rather than manual judgement, reducing error risk in stressful environments. Once connected to live traffic feeds, IoT sensors, and municipal traffic-signal infrastructure, CARRS could support automated signal pre-emption, dynamic corridor clearing for emergency vehicles, and coordination of multi-vehicle responses. By embedding legal and vehicle specific rules directly into routing logic, agencies minimize the risk of vehicles accessing unsuitable infrastructure (e.g. low tunnels, weight-restricted bridges), improving safety and compliance. Although full metropolitan graphs remain computationally heavy, the modular architecture supports future cloud-based distribution, hierarchical routing layers, and large-scale deployments across multiple cities or regions. Logs and constraint-violation records produced during routing could support post-incident review, infrastructure planning, and identification of recurrent congestion bottlenecks. In summary, if deployed, CARRS positions itself as a practical, high-impact application seeking to modernize emergency-response operations.

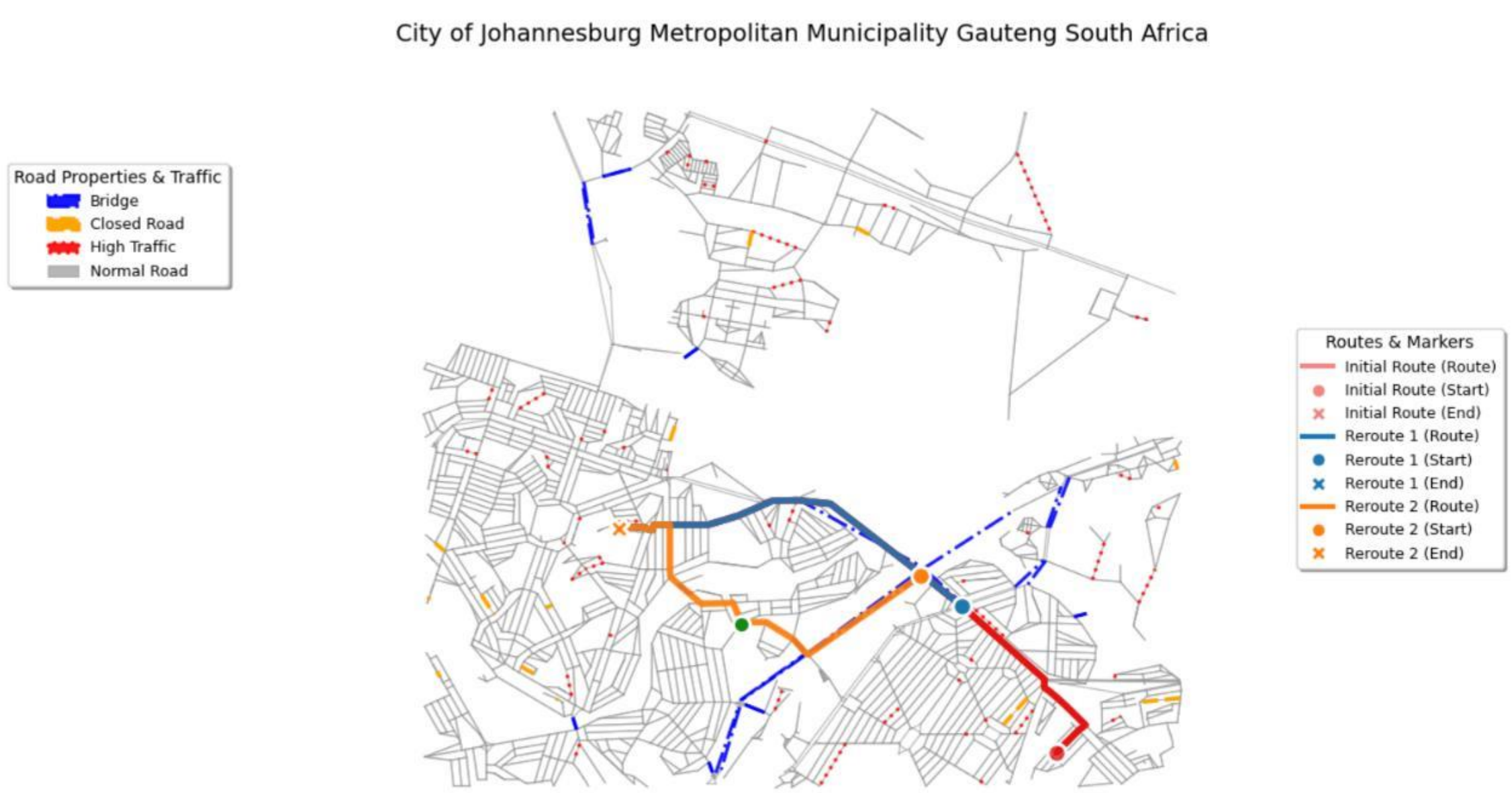

Figure 3: Dynamic rerouting visualization within the Johannesburg metropolitan area. The map illustrates the system's real-time adaptability, displaying the initial computed route (red) alongside subsequent consecutive reroutes (blue and orange) triggered by simulated traffic events and road closures

## 5 CONCLUSION

**Summary of Findings.** This study developed and validated the Constraint-Aware Real-Time Routing system, an artificial intelligence software engineering framework that integrates classical pathfinding algorithms with Constraint Satisfaction Problem (CSP) logic to provide dynamic, legally compliant, and computationally efficient routing for emergency vehicles in complex urban environments. Through extensive iterative simulation and evaluation, the system achieved 90% constraint adherence across vehicle types, route computation under 35 seconds (s) for operational subgraphs, rerouting responsiveness under 20s, responsive and user-friendly visualization supporting real-time coordination, and A* algorithm outperforming other algorithms in runtime and efficiency. The system thus demonstrates that real-time constraint-aware routing is both technically and operationally feasible for emergency services. By embedding constraint reasoning directly into the routing process, this system architecture bridges the gap between theoretical optimization and real-world emergency coordination.

**Limitations.** Despite robust performance, several imitations were identified, including the lack of scalability (full metropolitan graphs remain computationally expensive) and live data feeds, environmental constraints (weather, and dynamic incident data modelling), and potential infrastructure integration into live dispatch and traffic control systems.

**Future Research.** Building on this foundation, several research extensions are recommended to advance the capabilities and real-world applicability of the system. First, the integration with live data streams via APIs would incorporate real-time traffic, weather, and incident data from sources like smart sensors or transportation authorities. To manage the computational demands of such dynamic data at scale, the adoption of hierarchical and parallel routing architectures, using techniques like contraction hierarchies, or distributed computation is proposed to enable real-time routing across metropolitan areas. Furthermore, the development of predictive routing models that integrate machine learning would allow the system to anticipate congestion trends and pre-emptively re-route emergency vehicles.